\pdfoutput=1

\documentclass[11pt]{article}

\usepackage{acl}

\usepackage{times}
\usepackage{latexsym}
\usepackage{enumerate}
\usepackage{caption}
\usepackage{subcaption}
\usepackage{graphicx}
\usepackage{balance}

\usepackage[T1]{fontenc}

\usepackage[utf8]{inputenc}

\usepackage{microtype}

\usepackage{booktabs}

\title{Investigating data partitioning strategies \\ for crosslinguistic low-resource ASR evaluation}

\author{Zoey Liu \\
  Boston College \\
  \texttt{zoey.liu@bc.edu} \\\And
  Justin Spence \\
  University of California, Davis \\
  \texttt{jspence@ucdavis.edu} \\\And
Emily Prud'hommeaux \\
  Boston College \\
  \texttt{prudhome@bc.edu} \\}

\begin{document}
\maketitle
\begin{abstract}

Many automatic speech recognition (ASR) data sets include a single pre-defined test set consisting of one or more speakers whose speech never appears in the training set. This ``hold-speaker(s)-out" data partitioning strategy, however, 
may not be ideal for data sets in which the number of speakers is very small.
This study investigates ten different data split methods for five languages with minimal ASR training resources.
We find that (1) model performance varies greatly depending on which speaker is selected for testing;
(2) the average word error rate (WER) across all held-out speakers is comparable not only to the average WER over multiple random splits but also to any given individual random split;
(3) WER is also generally comparable when the data is split heuristically or adversarially;
(4) utterance duration and intensity are comparatively more predictive factors of variability regardless of the data split.
These results suggest that the widely used hold-speakers-out approach to ASR data partitioning can yield results that do not reflect model performance on unseen data or speakers. Random splits can yield more reliable and generalizable estimates when facing data sparsity.

\end{abstract}

\section{Introduction}

Certain model evaluation practices are considered standard or quite common in natural language processing (NLP), such as using popular benchmarks~\citep{bowman-etal-2015-large}, pre-defined data partitions~\citep{collins-2002-discriminative}, or random splits~\citep{gorman-bedrick-2019-need}.
All of these practices rely on metrics calculated over test sets as indices of model performance.
It is not generally acknowledged that a particular numerical result might be meaningful \textit{only for the specific train/test split that produced that result}. 
A single aggregated metric does not necessarily paint the full picture of a model architecture's potential~\citep{lewis-etal-2021-question}.

Automatic speech recognition (ASR) provides a case in point.
Given a data set produced by multiple speakers, the common data partitioning strategy is ``hold speaker(s) out", namely holding out all utterances from one or more speakers~\citep{panayotov2015librispeech,gauthier2016collecting}
as the test set, with the utterances from the remaining speakers serving as the training set.
Cross-validation is generally not applied; the speakers in the test set are fixed.
In other words, an ASR system is usually evaluated with just a single train/test split in which there is no speaker overlap between the training and test sets.

This common data partitioning strategy might fare well with a large data set, with recordings of dozens or hundreds of speakers, where the quantity of data and the wide array of speakers enable the training of models that are assumed to be speaker independent.
The same practice, however, is not ideal for low-resource scenarios, where the number of speakers is  much smaller. 
With endangered languages~\citep{meek2012we} in particular, there is much less flexibility in deciding how much data and what kind of utterances to include.
Thus, observed ASR accuracy may depend heavily on which speakers appear in the test sets rather than being representative of the model architecture's general performance. 

This study investigates alternative data partitioning methods for low-resource ASR.
Leveraging data from five typologically distinct languages, including one endangered language, we ask: 
(1) How dependent is ASR performance on the identity of the held-out speaker?
(2) Can alternative data partitioning strategies yield less variable estimates of a model's generalizability?
(3) What factors other than speaker identity contribute to differences in model performance?
(4) How can we operationalize lessons learned to improve ASR evaluation for under-resourced and endangered languages?


\section{Related Work}

While the hold-speaker(s)-out data partitioning method is prevalent in ASR~\citep{sikasote-anastasopoulos21bembaspeech,gauthier2016collecting,zeyer2019comparison,kipyatkova2016dnn}, there are a number of exceptions. 
Focusing on Wolof,~\citet{laleye2016FongbeASR} divided the utterances into three groups based on the categories of their content, then used two categories for training and one for testing.
~\citet{chiu2021rnn} tested the performance of an English ASR system trained on short audio segments with utterances of longer duration from YouTube and found poor generalization performance.
With five low-resource languages as the test cases,~\citet{ethan2021} re-partitioned the data where each speaker occurred in both the training and the test sets; the results showed considerable variability when compared to those derived from holding out one or a fixed set of speakers.

\section{Data descriptions}
\label{data}

We used data sets for four widely spoken low-resource languages, Fongbe~\citep{laleye2016FongbeASR}, Wolof~\citep{gauthier2016collecting}, Swahili~\citep{gelas:hal-00954048}, and Iban~\citep{Juan14}, which were previously released as ASR corpora. They include segmented audio with corresponding transcripts, as well as additional written texts for training the language model (Table~\ref{digitized} in Appendix~\ref{sec:descriptive_stats}).

In addition, we explored a data set of a critically endangered language indigenous to North America, which we refer to here as Language H. The audio recordings for Language H are the product of ongoing linguistic fieldwork started in 2005. All the recordings were produced by a single female elder speaker, which is common for speech corpora for critically endangered languages, making Language H a unique test bed for our study.
Each transcription typically goes through several stages of correction and consultation with the elder before being considered complete; thus some transcriptions have been examined more thoroughly than others.
Based only on differences in transcription quality, the audio data was divided into two sets, which we will call  ``verified"  vs. ``coarse" data  respectively (details are presented in Appendix~\ref{sec:hupa_diff}).

\section{Experiments}
\label{experiments}

\subsection{Data split methods}
\label{splits}

We first compared the commonly applied ``hold speaker(s) out" (hereafter \textbf{held-out speaker}) data partitioning strategy with random splits~\citep{gorman-bedrick-2019-need}.
For held-out speaker training, we set aside the data of one speaker for testing the performance of an acoustic model trained on the data of the other speakers. This procedure was repeated for all speakers in the data set.
Note that this data split method was only applicable to Wolof, Fongbe, and Iban.
For Swahili, which lacks information on speaker identity, and Language H, which includes only a single speaker, we adopted what we refer to as \textit{held-out session}.  Instead of holding out the data of each speaker, we held out the utterances from each recording date or fieldwork session.

For \textbf{random splits},
each data set was randomly divided into train/test sets so that the ratio between their respective total utterance duration approximated 4:1. To arrive at more reasonable comparisons with held-out speaker, for the data set(s) of each language, the number of random splits matched the number of speakers/sessions in total.

We also explored two alternative data splitting strategies: \textbf{heuristic} and \textbf{adversarial splits}~\citep{sogaard-etal-2021-need}.
For the former, we exploited the following features of each audio sample and its corresponding transcript: \textit{utterance duration}, \textit{average pitch}, \textit{average intensity}, \textit{the number of tokens} in the transcript, \textit{the number of unique token types} in the transcript, and \textit{the perplexity of the audio transcript} scored by the language model for each language (see Section~\ref{languagemodel}).
Consider the example of average pitch.
We identified a pitch threshold such that utterances with an average pitch value greater than or equal to this threshold would be put into the test set, and the total duration of these utterances accounted for around 20\% of the duration of the data set.
Note that each heuristic split method partitioned the data into a single train/test set split.

Lastly, for adversarial splits, we first combined the transcripts of all audio data for a particular data set, then split the transcripts into train/test sets via maximizing their Wasserstein distance~\citep{arjovsky2017wasserstein, sogaard-etal-2021-need}, so that the token distribution of utterances in the training set is as divergent or distant as possible from that of utterances in the test set. 
Each data set was split into train/test sets at a 4:1 ratio, five times.

\subsection{Language and acoustic models}
\label{languagemodel}
For each language, we used SRILM~\citep{stolcke2002srilm} to build a single trigram language model with Witten-Bell discounting using the additional written texts and excluding the transcripts of the audio training data.
For the acoustic model architecture, we used a fully connected deep neural network (DNN) from the open-source Kaldi toolkit~\citep{povey2011kaldi}, which is simple yet able to achieve effective performance in prior studies~\citep{ethan2021,georgescu2019kaldi,miao2015speaker}. 
In particular, for small corpora, the DNN architecture has been demonstrated to yield better results than statistical alternatives (e.g., subspace Gaussian mixture models) and other neural architectures (e.g., time delay neural networks)~\citep{ethanthesis}. We also found the DNN to be substantially more accurate than the endangered language end-to-end recipe~\citep{shi-etal-2021-leveraging} in ESPnet~\citep{watanabe2018espnet}. 
Crucially, however, we note that our goal is not to improve upon current state-of-the-art for low-resource ASR but rather to examine what data partitioning strategies and evaluation methods lead to reliable estimates in low-resource settings with an already very good model architecture.

\subsection{Regression analysis}
\label{regression}
To understand which features of the splits contribute to WER variability, we carried out regression analysis.
Given each data split, we first collected the following five heuristics
for \textit{each utterance} in the test set: utterance duration, average pitch, average intensity, utterance perplexity, and out-of-vocabulary (OOV) rate.
Second, we computed the \textit{average} value of each of the features for the training set as a whole.
Third,  we normalized the value of each feature for every utterance in the test set by the average value of the feature derived from the training set, to account for training set characteristics as well.
This yielded a reasonable data size for regression modeling for each language (6,248 instances for verified Language H to 85,920 instances for Wolof; see~\citet{anderson-etal-2021-replicating}).
After repeating these steps for all data splits, we 
fit regression models predicting the WER of every utterance in the test set as a function of these characteristics, while controlling for the number of tokens and types in the utterance and the data split method. When possible, speaker identity and the specific utterance were included as random effects, both with random intercept and slopes for each of the fixed effects. The final regression structure was determined via backward stepwise regression from the maximal mixed-effect  structure~\citep{barr2013random}.

\begin{figure*}[htb]
    \centering 
\begin{subfigure}{0.45\textwidth}
  \includegraphics[width=\linewidth]{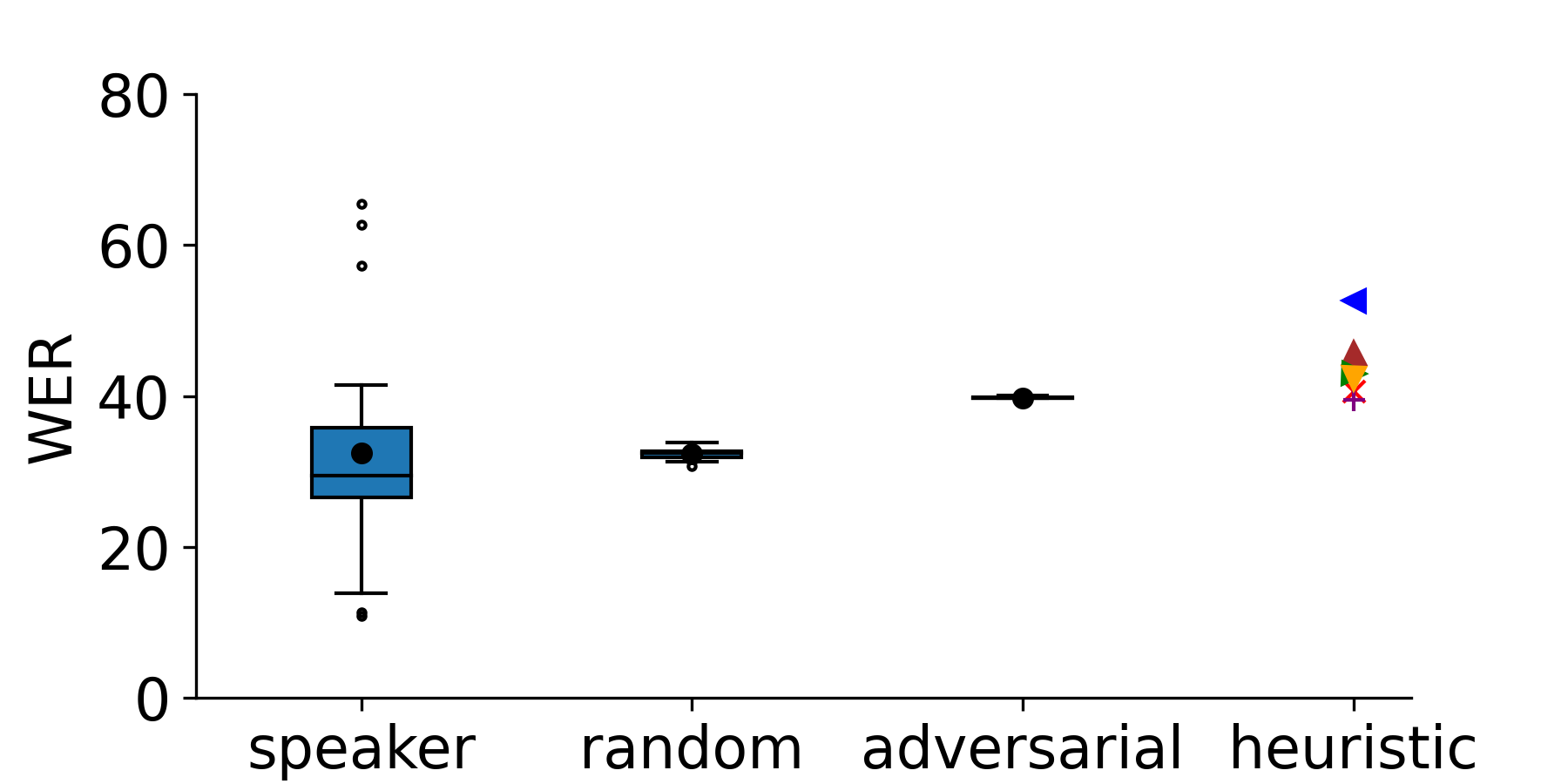}
  \caption{Fongbe}
  \label{fig:1}
\end{subfigure}\hfil 
\begin{subfigure}{0.45\textwidth}
  \includegraphics[width=\linewidth]{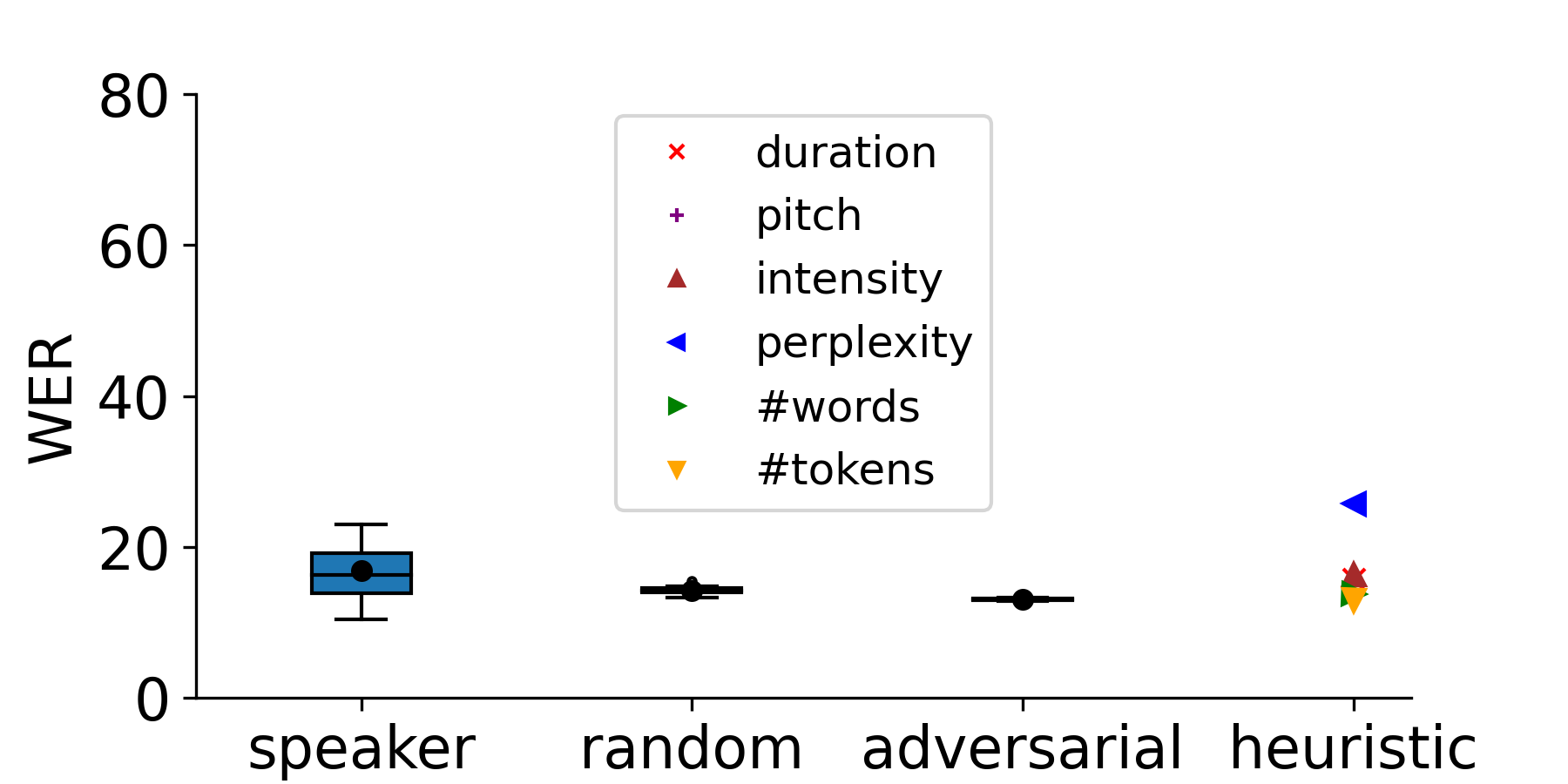}
  \caption{Iban}
  \label{fig:2}
\end{subfigure}\hfil 

\begin{subfigure}{0.45\textwidth}
  \includegraphics[width=\linewidth]{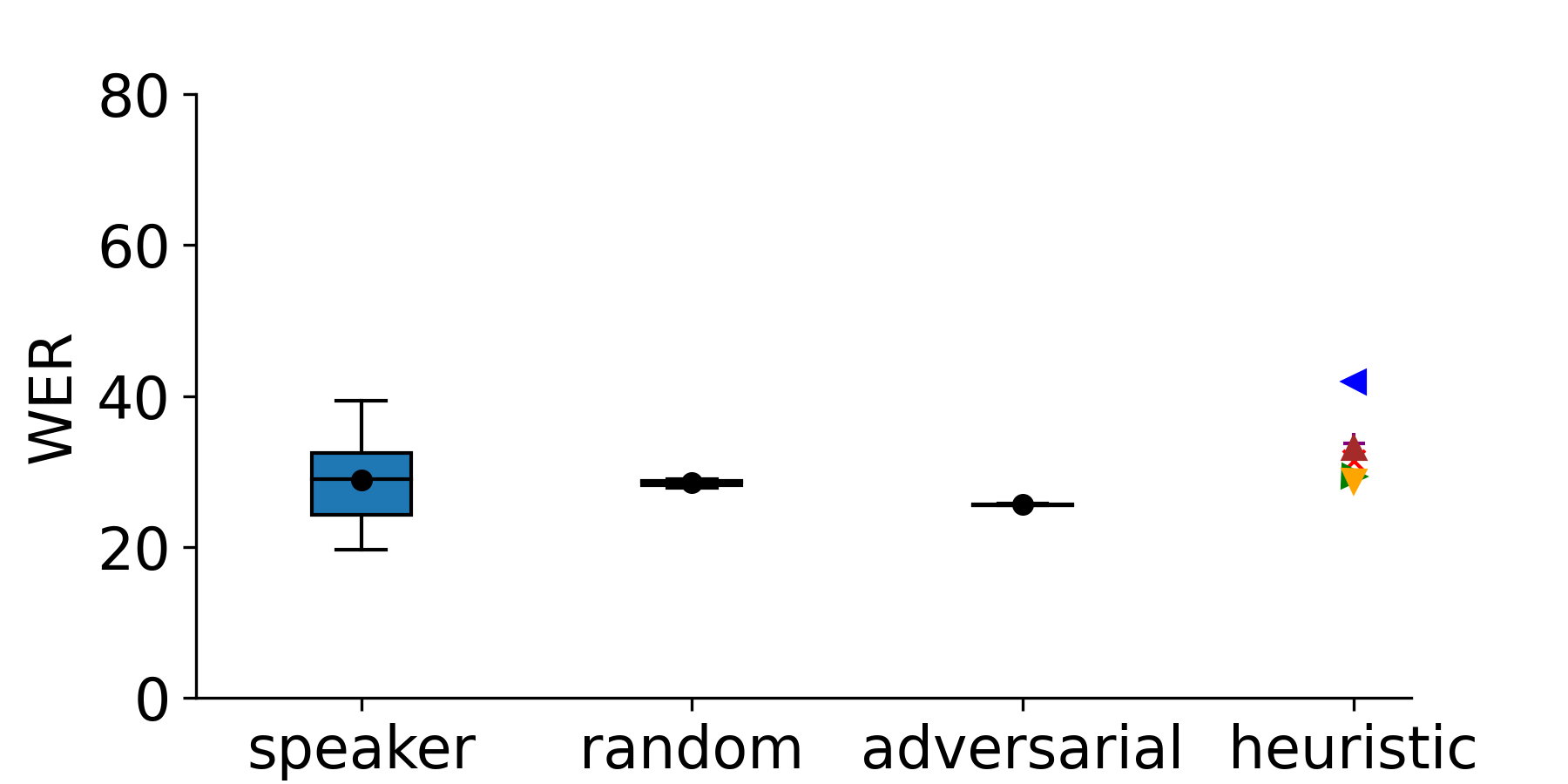}
  \caption{Wolof}
  \label{fig:3}
\end{subfigure}
\begin{subfigure}{0.45\textwidth}
  \includegraphics[width=\linewidth]{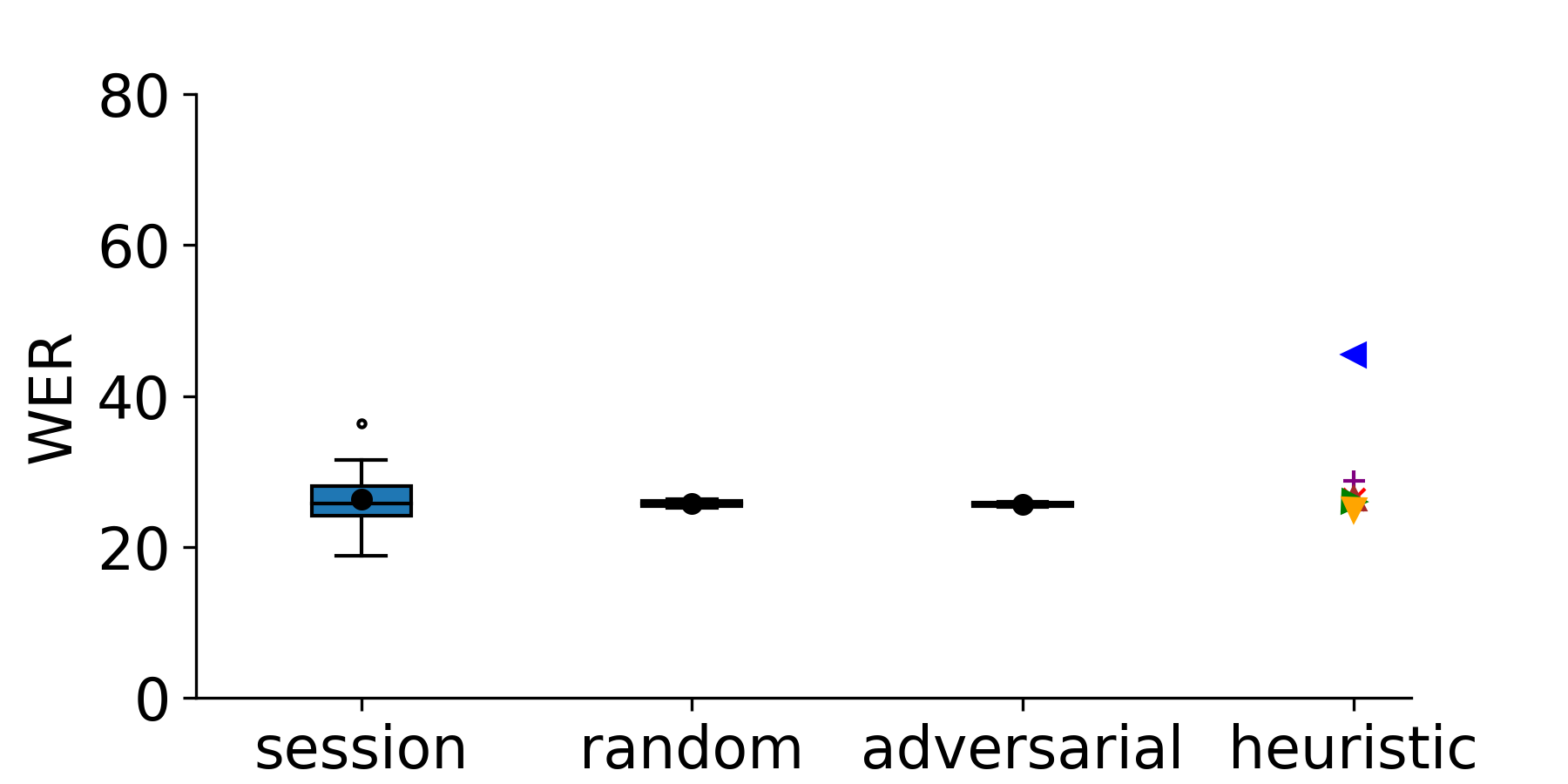}
  \caption{Swahili}
  \label{fig:4}
\end{subfigure}\hfil 

\begin{subfigure}{0.45\textwidth}
  \includegraphics[width=\linewidth]{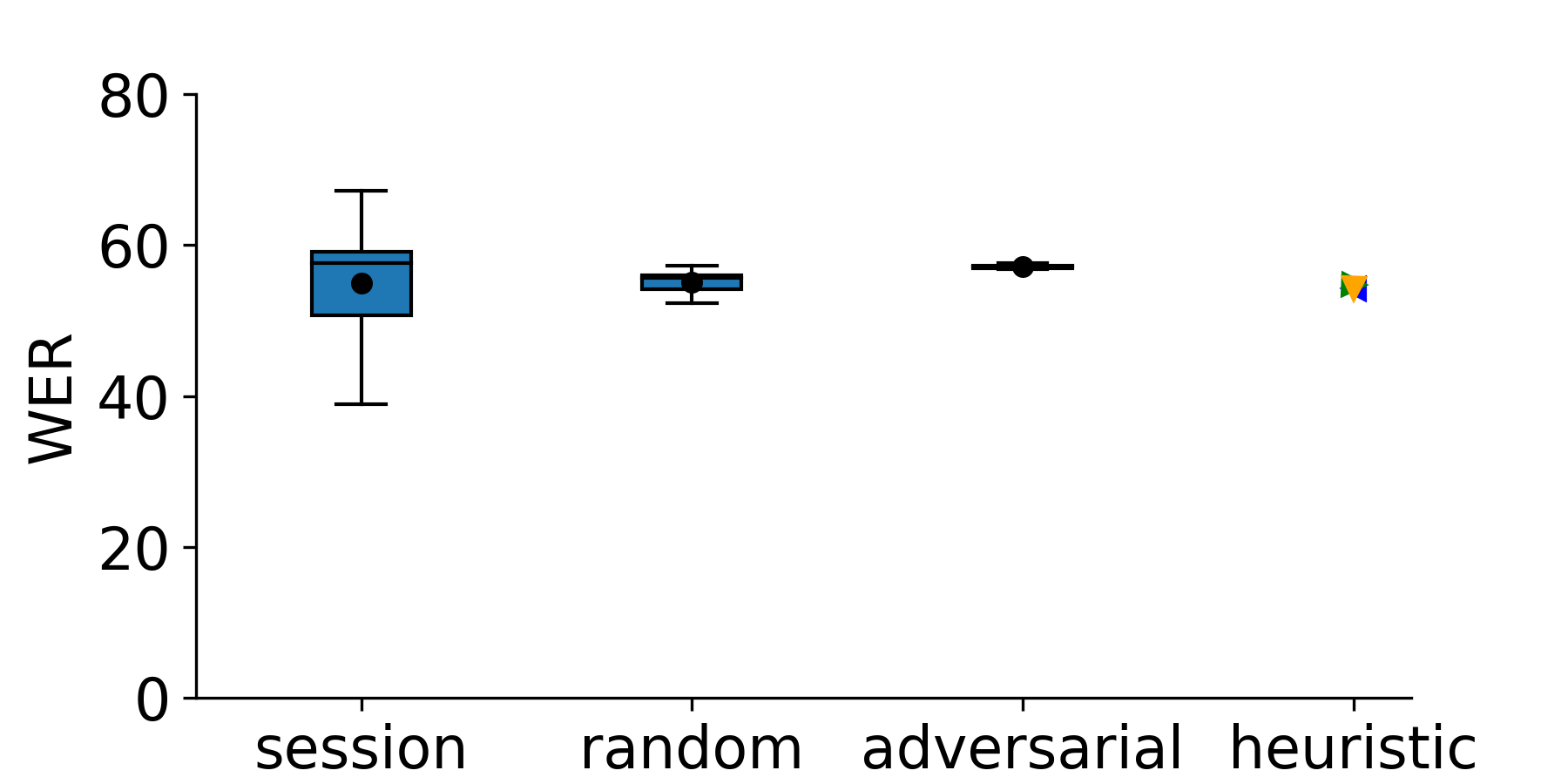}
  \caption{Language H (verified)}
  \label{fig:5}
\end{subfigure}\hfil 
\begin{subfigure}{0.45\textwidth}
  \includegraphics[width=\linewidth]{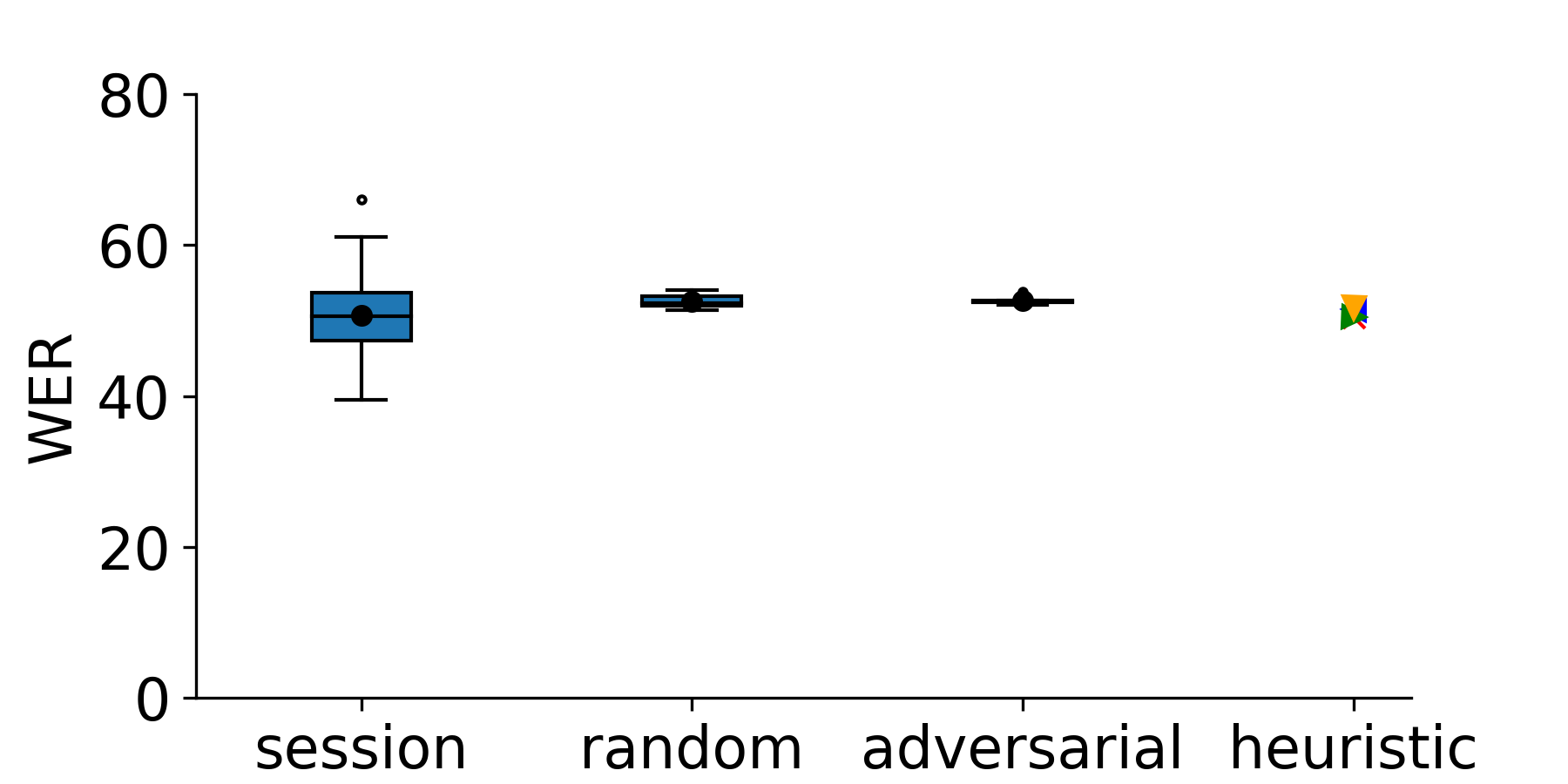}
  \caption{Language H (coarse)}
  \label{fig:6}
\end{subfigure}
\caption{WER for the various partitioning strategies for the six data sets. A large black dot represents the mean in each plot. Means for hold-session/speaker-out are comparable in all cases to means over random splits.}
\label{fig:images}
\end{figure*}

\section{Results}

We first consider the degree of variability in WER depending on which speaker is held out.
In Figure~\ref{fig:images}, for Fongbe, Wolof, and Iban, we see a high degree of variability.
The WER range across held-out speakers spans from 12.71 for Iban to 54.74 for Fongbe. (See Table~\ref{wer} in Appendix~\ref{sec:wer_results}.)
Perhaps surprisingly, for Swahili and Language H, where we held out recording sessions rather than individual speakers, we also observe great variability in model performance. 
The WER range across sessions is 17.59 for Swahili and is above 25 for both data sets of Language H.
Thus it does not appear to be the case that variability in WER is due entirely to the identity of the speaker; other factors such as recording setting or domain could contribute to this variability.
These observations speak to our original point, namely that observations from  ``hold speaker(s) out'' low-resource ASR evaluation are not representative of the model's generalizability.

One might suspect that the observed WER variability across speakers in each data set is (\textit{only}) because of the varying amount of audio available for each speaker.
Although there is a relationship between average WER and the total utterance duration per speaker (when looking at total utterance duration as a sole predictor in the regression) for Iban ($p < 0.005$), this relationship does not hold for Fongbe ($p = 0.75$) or Wolof ($p = 0.91$).
While there is a positive correlation between duration and WER for the coarse data of Language H ($p < 0.05$), this correlation does not exist for Swahili ($p = 0.45$) or for the verified data of Language H ($p = 0.99$). 
This indicates that the total amount of utterance duration \textit{alone} is not enough to yield (high) WER variability across speakers.

In contrast, results from random splits are much less variable. While not surprising, this is notable in that the average WER when holding out a speaker or session is comparable to that of random splits, or any one random split.
Thus a single random split can alone be enough to provide a reasonable estimate of the performance that would be derived by averaging over all random splits or over all possible held-out speakers/sessions.  In contrast, the WER for a model tested on single held-out speaker/session may not be a reliable estimate of the WER of that model on any other speaker.

On the other hand, splitting data heuristically and adversarially, creating test sets that consist of ``more challenging" cases than the training sets, does not necessarily lead to higher WER.
The results across the data split methods are mostly comparable except for when utterance duration or perplexity is used as the heuristic in certain cases.
Splitting the data by maximizing transcript distribution distance also yields minimal variability.


The regression analysis (Table~\ref{regression-table}, Appendix~\ref{sec:regression_results}) further indicates that most of the five features we investigated have significant effects on ASR model performance. Compared to other features, the ratios of utterance duration and intensity between the train and the test sets play strong roles more consistently  in predicting WER variability. 
The fact that utterance duration has an effect on WER \textit{when controlling for the effects of other factors} points to the potential limitation of evaluating models with held-out speakers in low-resource settings, where speakers contribute varying amounts of data.

\section{Discussion and Conclusion}

\looseness=-1 With data for four widely-spoken low-resource languages and one critically endangered language indigenous to North America,
our work demonstrates that there is a real risk of grossly over- or underestimating the performance of an ASR model architecture when evaluating on held-out speakers (and sessions) when only minimal resources are available. 
By contrast, random splits provide a more accurate and less variable estimate of the overall performance. 
Moreover, while cross-validation is advisable when partitioning by speaker, a single random split appears to provide an adequate estimate for expected WER on unseen data.


We note that these findings also hold for data sets partitioned according to recording session rather than speaker, suggesting that this phenomenon is not limited to diversity in speaker characteristics. This has implications particularly for ASR in support of endangered language documentation, in which the number of speakers is few but the recording conditions are highly variable.
We propose that future work on small ASR corpora for under-resourced languages carry out multiple evaluations on various data partitioning strategies in order to present a more complete picture of ASR model architecture performance. 



\bibliography{anthology,custom}
\bibliographystyle{acl_natbib}

\appendix

\section{Appendix}

\subsection{Descriptive statistics for the data of each language (Table~\ref{digitized})}
\label{sec:descriptive_stats}


\begin{table*}[ht!]
    \centering
    \footnotesize
    \begin{tabular}{llllllll}
        \toprule
       \multicolumn{1}{c}{\textbf{Language}} & \multicolumn{5}{c}{\textbf{Audio data}} & \multicolumn{2}{c}{\textbf{Additional written texts}} \\
        \cmidrule(lr){2-6} \cmidrule(lr){7-8} \\
        {} & $N$ of & Gender & Total utterance & Total utterance  & Total utterance  & $N$  & $N$  \\
  &  speakers/ & & duration per  & duration std.  & duration range  & of words & of types \\
        &  sessions & & speaker/session  &   &   &  &   \\
        \midrule
Fongbe & 27 & - & 16m12s & 7m12s & 22m12s & 990,146 & 8,022 \\
& &  &  &  & & \\
 Wolof & 18 & - & 63m & 18m36s & 1h6m36s & 601,639 & 29,147 \\
 & &  &  &  & & \\
Swahili & 36 & - & 18m & 7m12s & 19m12s & 31,540,821 & 471,296 \\
& &  &  &  & & \\
Iban & 23 & Male: 9 & 22m12s & 19m48s & 1h11m24s & 2,082,452 & 36,310 \\
& & Female: 14 & & & \\
& &  &  &  & & \\
Language H & 17 & Female: 1 & 5m24s & 6m & 22m48s &  41,386 & 8,800 \\
(verified) & &  &  &  & & \\
& &  &  &  & & \\
Language H & 34 & Female: 1 & 13m12s & 14m24s & 1h12s & & \\
(coarse) & & & & & & \\
        \bottomrule
    \end{tabular}
    \caption{
Descriptive statistics for audio data and additional written texts used to train language models for each language in the experiments; duration range refers to the range of the distribution of the total amounts of audio per speaker. We note that our counts were derived from the public repositories and may be different from those originally reported in the papers.
}
\label{digitized}
\end{table*}

\subsection{Data for Language H}
\label{sec:hupa_diff}

The majority of the recordings for Language H feature the elder telling stories from different genres, including traditional stories that explain how the world we know today came to be, personal anecdotes from her life, and oral-historical accounts of significant events in her speech community. 
Each recording has a time-aligned transcription produced by a human transcriber using annotation tools (e.g., ELAN~\citep{brugman2004annotating}); each transcript was rendered in a practical orthography currently adopted by the speech community.

The verified transcriptions for Language H are more accurate overall than coarse transcriptions and typically have undergone more orthographic normalization. 
This includes removing things that are audible in the recordings but not part of the standardized spelling (such as word-final epenthetic vowels), and removing false starts and other speech errors. 
In a small number of cases, verified transcriptions might even contain a word that is different from what was produced in the original recording, if the elder felt strongly that she had misspoken. 
Thus, although verified transcriptions tend to be more accurate than coarse ones, in some ways they are less faithful to the acoustic properties of the original recordings.

\subsection{Acoustic models}
\label{sec:acousticmodel}
Except for Swahili and Language H, acoustic feature transformations for the data of the other languages were conducted separately for each speaker. In detail, the recordings were transformed to the standard 13 dimensional mel-frequency cepstral coefficients (MFCCs), along with their delta- and delta-delta features. The delta- and delta-delta features are, respectively, numerical approximations of the first and second-order derivatives of the MFCCs and they were computed on a 25ms window with 10ms interval apart, allowing for modeling of the trajectories of the audio signals. Linear Discriminant Analysis and Maximum Likelihood Linear Transform were applied to reduce the dimensionality of the feature vectors. Speaker Adaptive Training was adopted to perform speaker and noise normalization in order to make the acoustic model more attentive to the phonemic variation present in the audio, rather than being restricted by the data of particular speakers. With the speaker-normalized features, Feature Space Maximum Likelihood Linear Regression (FMLLR) was employed for speaker-independent alignment.

The DNN we adopted had six hidden layers with 1024 units in each.
Sequence training was performed using the default parameters in Kaldi with state-level minimum Bayes risk criterion and a per-utterance Stochastic Gradient Descent weight update. Decoding was carried out with the Kaldi finite state transducer-based decoder.

\subsection{Regression results}
\label{sec:regression_results}

Full results for our regression analysis are presented in Table~\ref{regression-table}.

\begin{table*}[htb]
\footnotesize
\centering
\begin{tabular}{llllllll}
\hline
\textbf{Language} &  & \textbf{audio} & \textbf{Avg. pitch} & \textbf{Avg. intensity} & \textbf{utterance} & \textbf{OOV} & \textbf{$R^{2}$} \\
& & \textbf{duration} & & & \textbf{perplexity} & \\\hline
Fongbe & Coef. &  0.17*** & 0.13** & 0.11*** & 0.003*** & 0.008*** & 0.72 \\
& 95\% CI & (0.15, 0.20) & (0.11, 0.14) & (0.09, 0.13) & (0.001, 0.004) & (0.007, 0.009) \\
& & & & & & \\
Wolof & Coef. & 0.009 & 0.004 & 0.01 & 0.02*** & 0.00 & 0.73 \\
& 95\% CI & (-0.01, 0.03) & (-0.01, 0.02) & (-0.02, 0.04) & (0.02, 0.024) &  (0.00, 0.00)       \\
& & & & & & \\
Swahili & Coef. & 0.21*** & 0.06** & -1.65*** & 0.00 & 0.0131*** & 0.91 \\
& 95\% CI & (0.18, 0.24) & (0.02, 0.09) & (-1.81, -1.50) & (-6.30, 0.00) & (0.012, 0.014) \\
& & & & & & \\
Iban & Coef.  & 2.15*** & -0.10 & 0.58 & 0.00 & -0.002 & 0.99 \\
& 95\% CI & (1.87, 2.43) &  (-4.20, 0.21) &  (-2.16, 3.31) &  (-1.11, 0.00) & (-0.01, 0.01) \\
& & & & & & \\
Language H & Coef. & 1.30*** & -0.93*** & 2.74*** & 0.00 & 0.01*** & 0.95 \\
& 95\% CI &  (1.07, 1.52) &  (-1.13, -0.73) &  (2.09, 3.37) &  (-0.02, 0.01)  & (0.01, 0.02) &  \\
& & & & & & \\
Language H & Coef. & 0.13*** &  0.04 & -0.22** & -0.02*** & 0.01*** & 0.91 \\
& 95\% CI & (0.10, 0.16)  & (-0.01, 0.10) & (-0.35, -0.08) &  (-0.02, -0.01) & (0.01, 0.02) & \\
\hline
\end{tabular}
\caption{\label{regression-table}
Regression results for the data set(s) of each language in our experiments (CI stands for Confidence Interval); 
the number of * indicates significance level: * $p < 0.05$, ** $p < 0.01$, *** $p < 0.001$. Note that given the structure of our regression model, the coefficient value for the same feature is \textit{not comparable} across the data for each language (e.g., the coefficient of utterance duration ratio is 0.33 for the Wolof data, and 0.13 for the Swahili data; nevertheless, this does not mean that utterance duration ratio has a stronger effect for Wolof compared to its role for Swahili). 
Rather our goal is simply to see whether a feature potentially influences WER scores when the effects of other features are controlled for 
within the context of the data for every language.
}
\end{table*}

\subsection{Full WER results}
\label{sec:wer_results}
Table~\ref{wer} includes the full set of WER for every data partitioning strategy for each of the five languages.  which is represented visually in Figure~\ref{fig:images}.

\begin{table*}
\footnotesize
\centering
\begin{tabular}{ccccccccc}
\hline
\textbf{Language} & \textbf{Total} & \textbf{Split method} & \textbf{Threshold} & \textbf{$N$ of splits} & \textbf{WER} & \textbf{WER std.} &\textbf{WER range} \\
\hline
Fongbe & 7h10m & held-out speaker & - & 27 & 32.55 & 12.99 & 54.74 \\
& train:5h44m & random splits & - & 27 & 31.99 & 1.05 & 4.90 \\
& test:1h26m & utterance duration & 3.46s &  1 & 40.69 & - & - \\
& & Avg. pitch & 144.68hz & 1 & 38.90 & - & - \\
& & Avg. intensity & 58.24db & 1 & 43.00 & - & - \\
& & $N$ of tokens & 8 & 1 & 38.95 & - & - \\
& & $N$ of token types & 8 & 1 & 40.41 & - & - \\
& & utterance perplexity & 301.05 & 1 & 53.28 & - & - \\
& & distribution distance & - & 5 & 39.74 & 0.18 & 0.41 \\
\hline
Wolof & 18h58m & held-out speaker & - & 18 & 28.91 & 5.99 & 19.74 \\
& train:15h11m & random splits & - & 18 & 28.43 & 0.36 & 1.20 & \\
& test:3h47m & utterance duration & 5.19s & 1 & 31.37 & - & - \\
& & Avg. pitch & 114.43hz & 1 & 28.88 & - & - \\
& & Avg. intensity & 74.32db & 1 & 29.87 & - & - \\
& & $N$ of tokens & 11 & 1 & 28.07 & - & - \\
& & $N$ of token types & 11 & 1 & 28.85 & - & - \\
& & utterance perplexity & 674.09 & 1 & 42.63 & - & - \\
& & distribution distance & - & 5 & 25.65 & 0.08 & 0.19 \\
\hline
Swahili & 10h58m & held-out session & - & 36 & 26.31 &	3.46 &	17.59 \\
& train:8h47m & random splits & - & 36 & 25.83 & 0.45 & 2.09 \\
& test:2h11m & utterance duration & 4.80s & 1 & 26.36 & & \\
& & Avg. pitch & 172.07hz & 1 & 25.89 & - & - \\
& & Avg. intensity & 75.32db & 1 & 24.98 & - & - \\
& & $N$ of tokens & 13 & 1 & 25.36 & - & - \\
& & $N$ of token types & 12 & 1 & 25.37 \\
& & utterance perplexity & 1793.95 & 1 & 45.98 & - & - \\
& & distribution distance & - & 5 & 25.62 & 0.28 & 0.57 \\
\hline
Iban & 8h49m & held-out speaker & - & 23 & 16.92 & 3.80 & 12.71 \\
& train:6h49m & random splits & - & 23 & 14.35 & 0.57 & 2.19 \\
& test:1h42m & utterance duration & 15.85s & 1 & 15.73 & - & - \\
& & Avg. pitch & 141.33hz & 1 & 14.94 & - & - \\
& & Avg. intensity & 74.34db & 1 & 16.47 & - & - \\
& & $N$ of tokens & 36 & 1 & 13.97 & - & - \\
& & $N$ of token types & 31 & 1 & 13.90 & - & - \\
& & utterance perplexity & 361.55 & 1 & 27.02 & - & - \\
& & distribution distance & - & 5 & 13.09 & 0.14 & 0.36 & \\
\hline
Language H & 1h35m & held-out session & - & 17 & 55.73 & 8.59 & 31.82 \\
(verified) & train:1h16m & random splits & - & 17 & 55.27 & 1.50 & 5.44 \\
& test:19m & utterance duration & 9.71s & 1 & 60.12 & - & - \\
& & Avg. pitch & 112.40hz & 1 & 54.35 & - & - \\
& & Avg. intensity & 67.04db & 1 & 53.72 & - & - \\
& & $N$ of tokens & 16 & 1 & 52.17 & - & - \\
& & $N$ of token types & 14 & 1 & 54.43 & - & - \\
& & utterance perplexity & 898.45 & 1 & 53.55 & - & - \\
& & distribution distance & - & 5 & 55.10 & 0.47 & 1.28 \\
\hline
Language H & 7h37m & held-out session & - & 34 & 51.65 & 5.59 & 25.25 \\
(coarse) & train:6h6m & random splits & - & 34 & 52.85 & 1.18 & 4.63 \\
& test:1h31m & utterance duration & 10.96s & 1 & 56.20 & - & - \\
& & Avg. pitch & 113.78hz & 1 & 52.49 & & \\
& & Avg. intensity & 66.12db & 1 & 53.61 & & \\
& & $N$ of tokens & 16 & 1 & 50.36 & - & - \\
& & $N$ of token types & 14 & 1 & 50.76 & - & - \\
& & utterance perplexity & 933.37 & 1 & 50.26 & - & - \\
& & distribution distance & - & 5 & 51.39 & 0.18 & 0.47 \\
\hline
\end{tabular}
\caption{\label{wer}
WER results for the data set(s) of each language in our experiments; as we are focused on data partitioning strategy, for all data splits of a given data set, the language model was constant and was trained only on additional written texts. Note that one might be concerned about how much overlap there is between the test sets (and the training sets accordingly) yielded from different data partitioning strategies other than using held-out speaker/session; to address this, for the data set(s) of each language, we used the test set of the first random split as the reference and computed the proportion of overlapping utterances from the test sets of other data splits (except for held-out speaker/session); the maximum overlapping ratio across all the data sets was 0.25.}
\end{table*}

\end{document}